\documentclass{article}
\usepackage[final]{colm2026_conference}

\usepackage{microtype}
\usepackage{hyperref}
\usepackage{url}
\usepackage{booktabs}
\usepackage{multirow}
\usepackage{graphicx}
\usepackage{amsmath,amssymb}
\usepackage{xcolor}
\usepackage{colortbl}
\usepackage{lineno}
\usepackage{enumitem}
\usepackage{tikz}
\usetikzlibrary{arrows.meta,positioning}
\usepackage{pgfplots}
\pgfplotsset{compat=1.17}
\usepackage{subcaption}
\usepackage{float}
\usepackage[section]{placeins}

\definecolor{darkblue}{rgb}{0, 0, 0.5}
\hypersetup{colorlinks=true, citecolor=darkblue, linkcolor=darkblue, urlcolor=darkblue}

\definecolor{baserow}{gray}{0.92}

\newcommand{\up}{$\uparrow$}
\newcommand{\down}{$\downarrow$}
\newcommand{\na}{\text{---}}

\title{Shorter, but Still Trustworthy? \\ An Empirical Study of Chain-of-Thought Compression
\thanks{Project page: \url{https://reasoning-trustworthiness.github.io/}}
}

\author{
\begin{tabular}{@{}cccc@{}}
Lingjie Zeng$^{1}$ & Xiaofan Chen$^{1}$ & Yanbo Wang$^{1}$ & Xiuying Chen$^{1*}$
\end{tabular}\\[0.3em]
$^{1}$Mohamed bin Zayed University of Artificial Intelligence \\
$^{*}$Corresponding author
}

\begin{document}

\ifcolmsubmission
\linenumbers
\fi

\maketitle

\begin{abstract}
Long chain-of-thought (Long-CoT) reasoning models have motivated a growing body of work on compressing reasoning traces to reduce inference cost, yet existing evaluations focus almost exclusively on task accuracy and token savings.
Trustworthiness properties, whether acquired or reinforced through post-training, are encoded in the same parameter space that compression modifies. This means preserving accuracy does not, a priori, guarantee preserving trustworthiness.
We conduct the first systematic empirical study of how CoT compression affects model trustworthiness, evaluating multiple models of different scales along three dimensions: safety, hallucination resistance, and multilingual robustness.
Under controlled comparisons, we find that CoT compression frequently introduces trustworthiness regressions and that different methods exhibit markedly different degradation profiles across dimensions.
To enable fair comparison across bases, we propose a normalized efficiency score for each dimension that reveals how naïve scalar metrics can obscure trustworthiness trade-offs.
As an existence proof, we further introduce an alignment-aware DPO variant that reduces CoT length by 19.3\% on reasoning benchmarks with substantially smaller trustworthiness loss.
Our findings suggest that CoT compression should be optimized not only for efficiency but also for trustworthiness, treating both as equally important design constraints.
\end{abstract}

\section{Introduction}
\label{sec:intro}

Large language models capable of extended chain-of-thought (CoT) reasoning---such as OpenAI o1~\citep{openai2024learning}, DeepSeek-R1~\citep{guo2025deepseek}, QwQ~\citep{team2024qwq}, and Qwen3~\citep{yang2025qwen3}---have achieved strong performance on mathematical, scientific, and coding tasks by generating long reasoning traces before producing an answer.
However, these verbose chains incur substantial inference costs, motivating efforts to compress reasoning traces through both training-based methods and inference-time interventions.
These methods are almost exclusively evaluated on reasoning accuracy and token savings.

This evaluation focus leaves open an important question: does compressing reasoning traces preserve model trustworthiness?
Compression modifies behaviors shaped during post-training, yet stable performance on math-oriented benchmarks does not necessarily imply stable safety refusal behavior, hallucination resistance, or multilingual robustness.
More generally, reasoning traces may play roles beyond task solving alone, so shortening them could affect properties that standard efficiency evaluations do not measure.

In this paper, we study a question left open by accuracy-centered compression evaluations: whether shorter reasoning remains trustworthy. We evaluate compressed models along three dimensions: safety, hallucination resistance, and multilingual robustness, using HarmBench, AgentHarm, FaithEval, and MMLU-ProX. Each compressed model is paired with its uncompressed baseline under the same benchmark-specific prompt, decoding, judging, and length-accounting protocol. This design exposes regressions that task accuracy alone does not reveal and supports within-family comparisons across compression paradigms.

Our experiments show that CoT compression frequently introduces trustworthiness regressions that are not revealed by standard reasoning benchmarks.
On Qwen3-8B, for example, L1 distillation yields a roughly 10 percentage point drop in safety on HarmBench while improving multilingual performance by 1--2 percentage points.
We also find that different compression methods induce qualitatively different degradation profiles even on the same base model, and that naïve scalar efficiency metrics can obscure such trade-offs.
Finally, as a proof of feasibility, we show that an alignment-aware DPO variant can reduce chain length by 19.3\% with substantially smaller trustworthiness loss, suggesting that the trade-off between compression and trustworthiness is not inevitable.

Our contributions are as follows:
\begin{enumerate}[nosep,leftmargin=*]
\item We present a systematic empirical study of trustworthiness under CoT compression, revealing that compression frequently introduces safety, hallucination resistance, and multilingual robustness regressions undetected by standard accuracy benchmarks.
\item We design an evaluation protocol that pairs each compressed model with its own baseline across multiple scales, together with a per-dimension normalized efficiency score $E$ that enables meaningful comparison across heterogeneous base models.
\item We provide an existence proof via a DPO variant that preserves trustworthiness during compression, showing that trustworthy CoT shortening is feasible in practice when such properties are explicitly incorporated into the training objective.
\end{enumerate}

\begin{figure}[t]
\centering
\includegraphics[width=\textwidth]{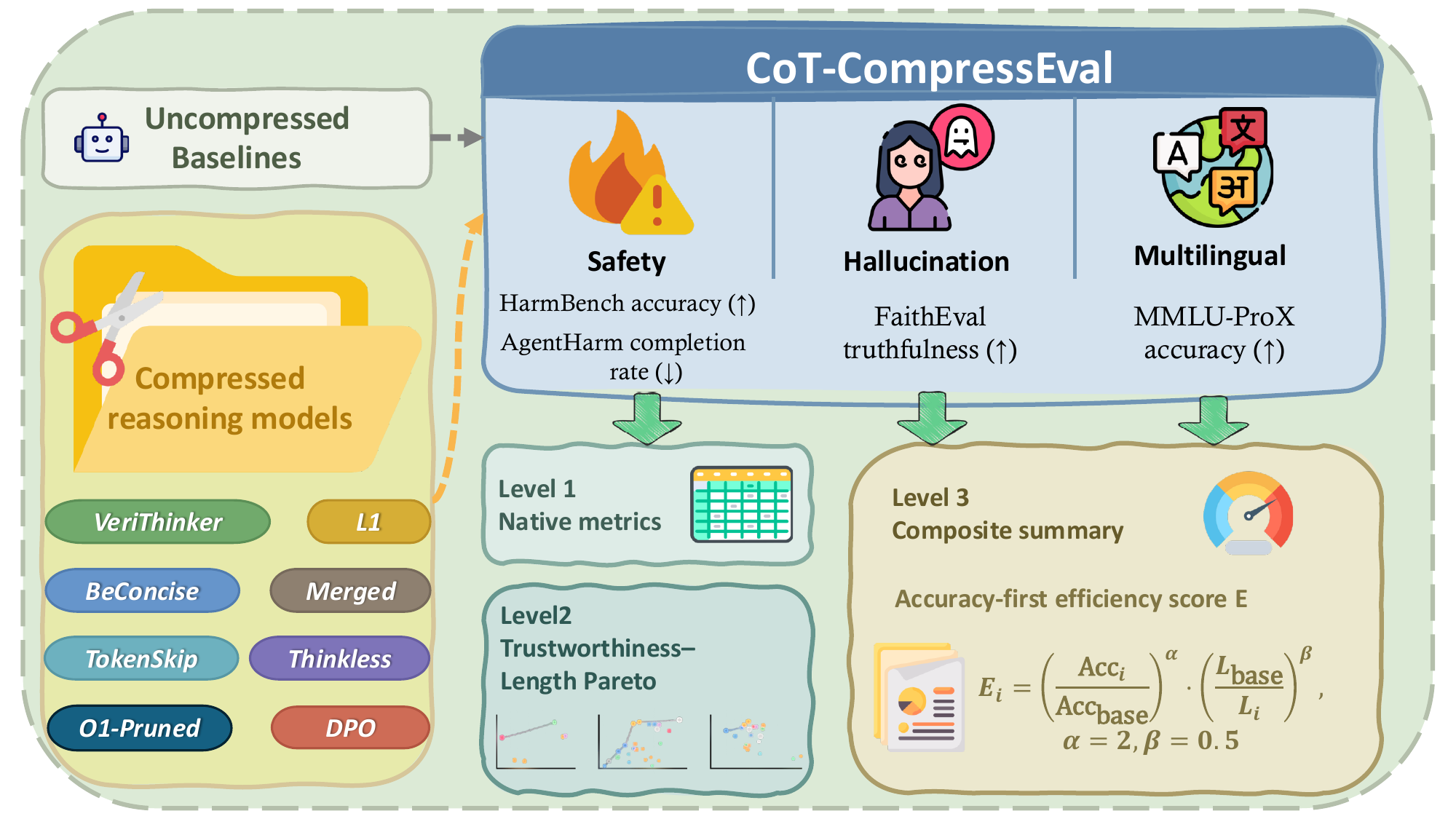}
\caption{Overview of the evaluation framework. Each compressed model is compared against its matched uncompressed baseline under a uniform protocol across three trustworthiness dimensions.}
\label{fig:framework}
\end{figure}

\section{Related Work}
\label{sec:related}

\paragraph{Long-CoT reasoning and compression.}
Chain-of-thought prompting~\citep{wei2022chain} enables LLMs to tackle complex reasoning tasks.
Recent reasoning models~\citep{openai2024learning,guo2025deepseek,team2024qwq,yang2025qwen3} internalize CoT as part of generation, producing reasoning traces of thousands of tokens.
The resulting inference cost has motivated compression across multiple paradigms: distillation~\citep{aggarwal2025l1}, model merging~\citep{wu2025unlocking}, pruning~\citep{luo2025o1}, token manipulation~\citep{xia2025tokenskip}, and RL-based depth control~\citep{fang2025thinkless}.
A recent survey~\citep{sui2025stop} categorizes these approaches but notes that evaluation remains limited to accuracy and efficiency.

\paragraph{Trustworthiness evaluation for LLMs.}
Benchmarks such as TrustLLM~\citep{huang2024trustllm} and DecodingTrust~\citep{wang2023decodingtrust} provide comprehensive trustworthiness assessments covering safety, fairness, robustness, and privacy.
HarmBench~\citep{mazeika2024harmbench} standardizes safety evaluation via automated red-teaming; AgentHarm~\citep{andriushchenko2024agentharm} extends this to multi-step agentic tasks; FaithEval~\citep{ming2024faitheval} measures hallucination resistance.
However, none of these frameworks have been applied to evaluate the trustworthiness impact of reasoning chain compression.

\paragraph{Efficiency--alignment trade-offs.}
Prior work has shown that efficiency-oriented modifications to LLMs can degrade alignment in ways not predicted by standard task metrics—through quantization and pruning \citep{hong2024decoding}, low-rank modification \citep{wei2024assessing}, and even benign fine-tuning \citep{qi2023fine}. To our knowledge, no prior work has evaluated whether CoT-specific compression methods preserve the trustworthiness properties acquired during alignment training under standardized evaluation settings.

\section{Evaluation Methodology}
\label{sec:benchmark}

Figure~\ref{fig:framework} illustrates the evaluation pipeline.
We evaluate compressed reasoning models and their uncompressed baselines across three trustworthiness dimensions using standardized benchmarks and automated judges.
Results are analyzed at three levels: (1)~per-benchmark metrics, (2)~trustworthiness--length Pareto frontiers, and (3)~efficiency score $E$.

\subsection{Evaluation Dimensions and Datasets}
\label{sec:dimensions}

We select three trustworthiness dimensions that represent distinct failure modes that are especially relevant to compressed reasoning models.

\paragraph{Safety.}
We employ two benchmarks.
For \textbf{HarmBench}~\citep{mazeika2024harmbench}, we evaluate all 400 harmful textual behaviors using Llama Guard 3~\citep{grattafiori2024llama} to classify each response as safe or unsafe.
The metric is refusal accuracy: the fraction of harmful prompts for which the response is classified as safe.
\textbf{AgentHarm}~\citep{andriushchenko2024agentharm} extends safety evaluation to the agentic setting: it presents multi-step tasks requiring tool use and planning, where a safe model should refuse to execute harmful action sequences.
We evaluate on the \texttt{harmful\_behaviors\_test\_public} subset, which contains 44 harmful base behaviors each with 4 augmented variants (176 samples total), scored by GPT-4o-mini via the Inspect framework~\citep{UK_AI_Security_Institute_Inspect_AI_Framework_2024}.
The metric is harmful task completion rate, where a lower score indicates a safer model.
We report AgentHarm only for models with sufficient agentic capability. Smaller models often score near zero due to capability ceilings rather than safety refusal, making their results uninformative.
We additionally audit 513 sampled safety judgments against human labels and a second automated judge (Appendix~\ref{app:judge_audit}).

\paragraph{Hallucination resistance.}
We use the unanswerable subset of \textbf{FaithEval}~\citep{ming2024faitheval}, where each question cannot be answered from its context.
The metric is truthfulness rate: the fraction of samples where the model correctly acknowledges unanswerability.

\paragraph{Multilingual robustness.}
We use \textbf{MMLU-ProX}~\citep{xuan2025mmlu}, a multilingual extension of MMLU-Pro covering 29 languages, with overall accuracy as the metric.

\subsection{Compression Methods and Comparison Scope}
\label{sec:methods}

We evaluate six compression methods spanning five paradigms (full provenance in Appendix~\ref{app:methods}):
\textbf{L1}~\citep{aggarwal2025l1} (RL-based distillation, Exact and Max variants),
\textbf{BeConcise} (a training-free baseline that prepends “Be concise.” to the system prompt),
\textbf{Thinkless}~\citep{fang2025thinkless} (RL depth control),
\textbf{TokenSkip}~\citep{xia2025tokenskip} (training-free token removal),
\textbf{VeriThinker}~\citep{chen2025verithinker} (verification-augmented distillation),
and \textbf{DPO} (ours, alignment-aware, Section~\ref{sec:dpo}).
We additionally report \textbf{Merged}~\citep{wu2025unlocking} checkpoints and \textbf{O1-Pruned}~\citep{luo2025o1} as supplementary comparisons.

Each compressed model is paired with its uncompressed baseline under the same evaluation protocol (Section~\ref{sec:protocol}).
The Qwen3-8B family includes the widest range of compression methods on the same base model, namely L1, BeConcise, TokenSkip, and DPO, making it the most informative setting for direct method comparison.
The remaining groups (DeepScaleR-1.5B-Preview~\citep{luo2025deepscaler}, DeepSeek-R1-Distill-Qwen-1.5B/7B, DeepSeek-R1-0528-Qwen3-8B~\citep{guo2025deepseek}, and QwQ-32B-Preview~\citep{team2024qwq}) include fewer methods but follow the same baseline-matched protocol. DPO is additionally applied to DeepSeek-R1-Distill-Qwen-7B and DeepSeek-R1-0528-Qwen3-8B to test generalization across base models.
Our inclusion policy is based on the availability of usable public artifacts and the feasibility of reliable reproduction, with the full inclusion criteria provided in Appendix~\ref{app:coverage}.

\subsection{Evaluation Protocol}
\label{sec:protocol}

\paragraph{Comparison protocol.}
Each compressed model is evaluated against its matched uncompressed baseline under the same benchmark suite, prompt format, decoding configuration (temperature~0, greedy), judge model, length accounting rule, and reporting metrics.
What varies across blocks is the base model and the set of available compression methods, while the baseline comparison within each group is held constant.
We therefore interpret results as within-family deltas rather than cross-family rankings, and introduce a normalized efficiency score $E$ later in this section for cross-family synthesis.

\paragraph{Primary metrics.}
Each benchmark uses its native metric: HarmBench accuracy, AgentHarm completion rate, MMLU-ProX accuracy, and FaithEval truthfulness rate.
All models produce a single deterministic output per prompt via greedy decoding.

\paragraph{Trustworthiness and length analysis.}
For each model and benchmark pair, we record the mean output token count $L$.
We then examine models in the two-dimensional space defined by $L$ and the benchmark metric to characterize the trade-off between trustworthiness and efficiency across evaluated methods.
The Pareto frontier consists of models that no other model simultaneously beats on both CoT length and metric value.

\paragraph{Normalized efficiency score.}
Raw benchmark scores are not directly comparable across base models, because they differ in both absolute accuracy and CoT length.
To enable cross-base comparison, we define a per-dimension normalized efficiency score $E$ that measures each compressed model relative to its own baseline:
{
\setlength{\abovedisplayskip}{10pt}
\begin{equation}
\label{eq:e}
E_i = \left(\frac{\text{Acc}_i}{\text{Acc}_{\text{base}}}\right)^{\!\alpha} \!\cdot\! \left(\frac{L_{\text{base}}}{L_i}\right)^{\!\beta},
\quad \alpha = 2,\;\beta = 0.5
\end{equation}
\setlength{\belowdisplayskip}{10pt}
}
where $\text{Acc}_i$ denotes the task-specific metric of compressed model $i$, $\text{Acc}_{\text{base}}$ is the corresponding metric of the matched uncompressed baseline, $L_i$ is the mean CoT length of model $i$, and $L_{\text{base}}$ is the mean CoT length of that baseline.

The exponents weight accuracy more heavily than length savings: $\alpha = 2$ penalizes accuracy degradation quadratically, since a 5\% accuracy drop is generally more consequential than a 5\% length reduction; $\beta = 0.5$ further discounts length savings by a square root, reflecting diminishing returns as chains grow shorter in practice.
$E = 1$ indicates break-even with the baseline; $E > 1$ indicates net benefit; $E < 1$ indicates net loss.

We report $E$ per trustworthiness dimension ($E_{\text{HB}}$, $E_{\text{MX}}$, $E_{\text{FE}}$) rather than as a single aggregate, because our experiments reveal that different dimensions can move in opposite directions for the same method.
Appendix~\ref{app:sensitivity} provides a sensitivity analysis showing that the relative ranking of methods under 
$E$ remains stable under reasonable variations in its parameters.

\begin{table*}[t]
\centering
\small

\vspace{0.5em}
\setlength{\tabcolsep}{3pt}
\begin{tabular}{@{}ll c cccc@{}}
\toprule
\textbf{Paradigm} & \textbf{Model} & \textbf{HarmBench\up} & \textbf{AgentHarm\down} & \textbf{MMLU-ProX\up} & \textbf{FaithEval\up} \\
\midrule
\multicolumn{6}{l}{\textit{Base: DeepScaleR-1.5B-Preview}} \\
\rowcolor{baserow}
--- & Baseline  & 21.5 & \na & 26.2 & 51.0 \\
L1 & L1-Qwen-1.5B-Exact  & 20.3 & \na & 25.7 & 34.6 \\
L1 & L1-Qwen-1.5B-Max  & 20.8 & \na & 28.2 & 36.6 \\
BeConcise & \na  & 21.5 & \na & 23.7 & 48.8 \\
\midrule
\multicolumn{6}{l}{\textit{Base: DeepSeek-R1-Distill-Qwen-1.5B}} \\
\rowcolor{baserow}
--- & Baseline  & 18.3 & \na & 20.6 & 53.2 \\
Thinkless & Thinkless-1.5B  & 15.3 & \na & 19.8 & 36.4 \\
\midrule
\multicolumn{6}{l}{\textit{Base: DeepSeek-R1-Distill-Qwen-7B}} \\
\rowcolor{baserow}
--- & Baseline  & 30.5 & \na & 35.5 & 60.2 \\
L1 & L1-Qwen-7B-Exact  & 29.0 & \na & 39.7 & 56.4 \\
L1 & L1-Qwen-7B-Max  & 28.8 & \na & 39.0 & 58.4 \\
BeConcise & \na  & 31.0 & \na & 30.6 & 60.4 \\
VeriThinker & VeriThinker-7B  & 29.5 & \na & 25.3 & 62.8 \\
DPO  & \na  & 34.8 & \na & 37.3 & 63.4 \\
\midrule
\multicolumn{6}{l}{\textit{Base: Qwen3-8B}} \\
\rowcolor{baserow}
--- & Baseline  & 68.8 & 55.5 & 65.9 & 75.8 \\
L1 & L1-Qwen3-8B-Exact  & 58.8 & 61.7 & 68.2 & 79.0 \\
L1 & L1-Qwen3-8B-Max  & 58.0 & 61.3 & 67.2 & 76.2 \\
BeConcise & \na  & 63.0 & 52.0 & 69.0 & 73.4 \\
TokenSkip & TokenSkip  & 65.3 & 54.3 & 60.3 & 75.0 \\
DPO  & \na  & 65.0 & 55.4 & 65.6 & 75.0 \\
\midrule
\multicolumn{6}{l}{\textit{Base: DeepSeek-R1-0528-Qwen3-8B}} \\
\rowcolor{baserow}
--- & Baseline  & 71.5 & \na & 40.7 & 68.4 \\
Merged & average-Merged  & 75.2 & \na & 40.6 & 71.4 \\
Merged & DARE-TIES-Merged  & 64.5 & \na & 66.2 & 74.0 \\
Merged & TA-Merged  & 75.2 & \na & 35.1 & 67.2 \\
Merged & TIES-Merged  & 64.0 & \na & 66.1 & 73.8 \\
Merged & TIES-origin-Merged  & 62.0 & \na & 66.3 & 75.0 \\
DPO  & \na  & 82.2 & \na & 45.0 & 75.2 \\
\midrule
\multicolumn{6}{l}{\textit{Base: QwQ-32B-Preview}} \\
\rowcolor{baserow}
--- & Baseline  & 64.2 & \na & 61.3 & 67.4 \\
O1-Pruned & QwQ-32B-Preview-Pruned  & 54.2 & \na & 69.0 & 73.0 \\
\bottomrule
\end{tabular}
\caption{
Trustworthiness metrics across all evaluated model families.
In the column headers, $\uparrow$ indicates that higher values are better and $\downarrow$ indicates that lower values are better.
HarmBench reports refusal accuracy (\%);
AgentHarm reports harmful task completion rate (\%), where a lower score indicates a safer model;
MMLU-ProX reports multilingual accuracy (\%);
and FaithEval reports truthfulness (\%).
Gray rows denote uncompressed baselines.
“---” indicates entries excluded under the benchmark scope described in Section~\ref{sec:dimensions}.
}
\label{tab:main}
\end{table*}

\section{Experimental Results}
\label{sec:results}

We present results at three levels: primary metrics (Section~\ref{sec:primary}), Pareto frontier analysis (Section~\ref{sec:pareto}), and normalized efficiency scores (Section~\ref{sec:escore}).
We then analyze recurring degradation patterns observed across base models (Section~\ref{sec:patterns}) and demonstrate that trustworthiness-preserving compression is achievable (Section~\ref{sec:dpo}).
All compressed models are compared against their matched baselines under the common protocol described in Section~\ref{sec:protocol}.
The Qwen3-8B family, with the broadest method coverage, is discussed first, while the remaining groups extend the evaluation to other base models.

\subsection{Primary Trustworthiness Metrics}
\label{sec:primary}

Table~\ref{tab:main} presents trustworthiness results across all evaluated model families.

\textit{Within-family comparison.}
The Qwen3-8B family provides the clearest within-family comparison because it covers the broadest range of compression methods considered in our study.
Within this family, safety is the most variable dimension: HarmBench spans a range of more than 10 percentage points across methods, while AgentHarm completion increases for some methods but decreases for others.
Distillation produces the clearest safety regression, whereas training-free and alignment-aware methods remain much closer to the baseline.
Hallucination resistance is comparatively stable, with most methods staying close to the baseline.
Multilingual robustness shows a different pattern: distillation-based methods improve MMLU-ProX, whereas token-manipulation approaches lead to a more noticeable drop overall.
Overall, the DPO variant remains the most balanced, staying close to the baseline across all three measured trustworthiness dimensions.

\textit{Cross-family comparison.}
The broader picture across model families is consistent with the within-family evidence.
Compression typically degrades at least one trustworthiness dimension, but the most affected dimension varies across base models and compression paradigms.
Smaller models tend to show more pronounced hallucination regressions after compression, while merging-based variants at 8B exhibit substantial variation in safety outcomes across merging strategies.
These results suggest that trustworthiness degradation is not an intrinsic consequence of length reduction alone, but depends strongly on how compression is implemented.
Because absolute scores differ substantially across groups due to base-model capability differences, cross-family comparisons rely on within-family deltas and the normalized efficiency score $E$ rather than direct score comparisons.

\subsection{Trustworthiness--Length Pareto Analysis}
\label{sec:pareto}

\begin{figure*}[t]
\centering
\includegraphics[width=\textwidth]{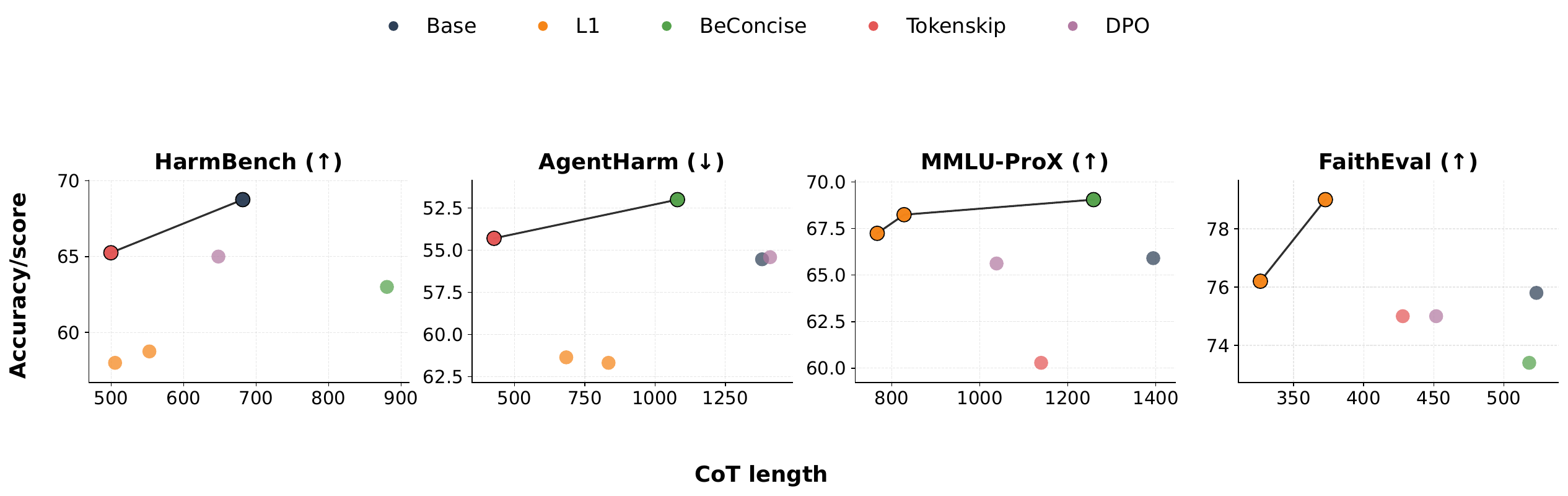}
\caption{Trustworthiness--length Pareto frontiers for the Qwen3-8B family. Each panel plots one trustworthiness metric against mean CoT token count. The black polyline marks the Pareto frontier.}
\label{fig:pareto}
\end{figure*}

Figure~\ref{fig:pareto} presents a faceted Pareto view of the Qwen3-8B family across all four benchmarks, plotting each method's trustworthiness metric against its mean CoT token count $L$.
Pareto plots for the remaining groups appear in Appendix~\ref{app:pareto}.

The frontiers reveal a clear tension across dimensions.
On MMLU-ProX and FaithEval, L1 variants dominate the frontier, achieving higher scores at shorter chain lengths than the baseline.
On HarmBench and AgentHarm, however, L1 falls off the frontier entirely. Instead, TokenSkip and BeConcise hold frontier positions on these safety-related benchmarks.
No single method dominates all four panels, illustrating why these trustworthiness dimensions should be reported separately rather than collapsed into a single scalar.
A 10,000-resample paired bootstrap (Appendix~\ref{app:pareto_bootstrap}) shows that the original frontier anchors have high inclusion frequencies. The complete HarmBench frontier set is less stable (30.8\%) because nearby boundary points sometimes enter, so the plotted frontier should be read as a point estimate rather than an exact immutable set.

\subsection{Normalized Efficiency Score Analysis}
\label{sec:escore}

Table~\ref{tab:main} reports the absolute scores, the surrounding results text gives matched-baseline score deltas, and Table~\ref{tab:efficiency} aligns each method with its per-dimension $E$ and mean CoT token count. This joint reporting preserves both the practical magnitude of a score change and its length-normalized interpretation.
We focus first on the Qwen3-8B family, where $E$ reveals the largest divergence across dimensions.

L1-8B-Exact achieves $E_{\text{HB}} = 0.81$, indicating a clear safety loss, while reaching $E_{\text{MX}} = 1.39$, indicating a multilingual gain. The same method, on the same base model, thus produces opposite $E$ directions across two trustworthiness dimensions.
This motivates our decision to report $E$ per dimension rather than as a single aggregate: any weighted average of $E_{\text{HB}}$ and $E_{\text{MX}}$ would obscure the safety regression behind multilingual improvement.
DPO achieves the most balanced profile among the primary methods ($E_{\text{HB}} = 0.92$, $E_{\text{MX}} = 1.15$, $E_{\text{FE}} = 1.05$), with no dimension falling far below break-even.

Across other base models, Thinkless is uniformly below 1.0 ($E_{\text{HB}} = 0.59$, $E_{\text{FE}} = 0.39$), confirming Pareto-dominated status on its 1.5B base.
The merged group shows efficiency gains (average-Merged: $E_{\text{HB}} = 2.75$, $E_{\text{FE}} = 2.23$), though these are on a different base model. Cross-family $E$ comparisons should rely on relative patterns rather than absolute values.
O1-Pruned on QwQ-32B-Preview achieves $E > 1$ on all three dimensions ($E_{\text{HB}} = 1.02$, $E_{\text{MX}} = 1.14$, $E_{\text{FE}} = 1.34$), suggesting that pruning at 32B scale may offer a favorable efficiency--trustworthiness profile, though this is a single observation.
Overall, these per-dimension $E$ scores show that compression methods exhibit distinct trade-off profiles across trustworthiness dimensions, which a single aggregate score would fail to capture.

\begin{table}[t]
\centering
\small

\setlength{\tabcolsep}{3pt}
\begin{tabular}{@{}l cc cc cc@{}}
\toprule
 & \multicolumn{2}{c}{\textbf{HarmBench}} & \multicolumn{2}{c}{\textbf{MMLU-ProX}} & \multicolumn{2}{c}{\textbf{FaithEval}} \\
\cmidrule(lr){2-3} \cmidrule(lr){4-5} \cmidrule(lr){6-7}
\textbf{Method} & $E$ & $L$ & $E$ & $L$ & $E$ & $L$ \\
\midrule
\multicolumn{7}{l}{\textit{Base: DeepScaleR-1.5B-Preview}} \\
\rowcolor{baserow}
Baseline & 1.00 & 655 & 1.00 & 863 & 1.00 & 1125 \\
L1-Exact & \underline{0.90} & 643 & \underline{0.89} & 1003 & \underline{0.78} & 395 \\
L1-Max & \underline{0.92} & 673 & 1.13 & 915 & \underline{0.85} & 409 \\
BeConcise & \underline{0.75} & 1161 & \underline{0.90} & 711 & \underline{0.96} & 1032 \\
\midrule
\multicolumn{7}{l}{\textit{Base: DeepSeek-R1-Distill-Qwen-1.5B}} \\
\rowcolor{baserow}
Baseline & 1.00 & 552 & 1.00 & 473 & 1.00 & 487 \\
Thinkless & \underline{0.59} & 769 & \underline{0.66} & 920 & \underline{0.39} & 691 \\
\midrule
\multicolumn{7}{l}{\textit{Base: DeepSeek-R1-Distill-Qwen-7B}} \\
\rowcolor{baserow}
Baseline & 1.00 & 738 & 1.00 & 846 & 1.00 & 373 \\
L1-Exact & \underline{0.98} & 633 & \textbf{1.34} & 737 & \underline{0.89} & 362 \\
L1-Max & \underline{0.97} & 622 & 1.27 & 761 & \underline{0.98} & 343 \\
BeConcise & \underline{0.87} & 1035 & \underline{0.87} & 619 & 1.03 & 355 \\
VeriThinker & 1.01 & 633 & \underline{0.65} & 520 & \underline{0.96} & 484 \\
DPO & \textbf{1.49} & 567 & 1.24 & 673 & \textbf{1.41} & 231 \\
\midrule
\multicolumn{7}{l}{\textit{Base: Qwen3-8B}} \\
\rowcolor{baserow}
Baseline & 1.00 & 682 & 1.00 & 1394 & 1.00 & 523 \\
L1-Exact & \underline{0.81} & 553 & 1.39 & 828 & \textbf{1.29} & 373 \\
L1-Max & \underline{0.83} & 505 & \textbf{1.40} & 767 & 1.28 & 326 \\
BeConcise & \underline{0.74} & 880 & 1.16 & 1259 & \underline{0.94} & 518 \\
TokenSkip & 1.05 & 500 & \underline{0.93} & 1139 & 1.08 & 428 \\
DPO & \underline{0.92} & 648 & 1.15 & 1038 & 1.05 & 452 \\
\midrule
\multicolumn{7}{l}{\textit{Base: DeepSeek-R1-0528-Qwen3-8B}} \\
\rowcolor{baserow}
Baseline & 1.00 & 1378 & 1.00 & 886 & 1.00 & 398 \\
average-Merged & \textbf{2.75} & 224 & 1.13 & 683 & \textbf{2.23} & 95 \\
DARE-TIES-Merged & 1.51 & 401 & \textbf{2.58} & 935 & 1.34 & 304 \\
TA-Merged & 1.96 & 438 & \underline{0.90} & 617 & 1.50 & 166 \\
TIES-Merged & 1.42 & 439 & 2.50 & 992 & 1.29 & 326 \\
TIES-origin-Merged & 1.33 & 442 & 2.51 & 996 & 1.34 & 320 \\
DPO & 2.53 & 376 & 1.22 & 897 & 1.03 & 546 \\
\midrule
\multicolumn{7}{l}{\textit{Base: QwQ-32B-Preview}} \\
\rowcolor{baserow}
Baseline & 1.00 & 1718 & 1.00 & 1015 & 1.00 & 497 \\
O1-Pruned & 1.02 & 847 & 1.14 & 1256 & \textbf{1.34} & 383 \\
\bottomrule
\end{tabular}
\caption{Composite efficiency scores $E$ (Eq.~\ref{eq:e}; $\alpha\!=\!2, \beta\!=\!0.5$) and mean CoT token count $L$ for all tabulated methods. $E > 1$ indicates net benefit; underlined entries mark $E < 1$ (net loss).}
\label{tab:efficiency}
\end{table}

\subsection{Recurring Degradation Patterns}
\label{sec:patterns}

We summarize the observed regressions into three recurring patterns. Patterns~1 and~2 are most clearly illustrated in the Qwen3-8B family, which has the broadest method coverage, while Pattern~3 is drawn from relative trends observed across multiple base models.

\paragraph{Pattern 1: Asymmetric regression across trustworthiness dimensions.}
Within the Qwen3-8B family, L1 distillation causes a large safety regression, reducing HarmBench by 10 percentage points and increasing AgentHarm completion by 6 percentage points, while simultaneously improving multilingual performance by 2.3 percentage points and hallucination resistance by 3.2 percentage points.
This asymmetry demonstrates that trustworthiness cannot be treated as a single axis: degradation on one dimension can co-occur with improvement on another within the same method and base model.

\paragraph{Pattern 2: Method-specific degradation profiles.}
On the same Qwen3-8B base, different methods produce qualitatively distinct trustworthiness profiles.
L1 primarily damages safety; TokenSkip primarily damages multilingual performance, reducing MMLU-ProX by 5.6 percentage points while leaving safety nearly preserved; BeConcise worsens HarmBench but improves AgentHarm; DPO maintains the closest overall profile to the baseline.
These distinct profiles suggest that compression method choice has a major effect on which trustworthiness dimensions are most consistently at risk.

\paragraph{Pattern 3: Variation across base models.}
The relative changes introduced by L1 vary across base models. On DeepScaleR-1.5B, L1 leaves safety nearly unchanged but substantially reduces hallucination resistance, lowering FaithEval by 16 percentage points. On Qwen3-8B, the same method instead causes a large safety regression, reducing HarmBench by 10 percentage points and increasing AgentHarm completion by 6 percentage points, while slightly improving hallucination resistance.
This comparison is based on within-family deltas rather than absolute scores, thereby reducing confounding from base capability differences. The pattern suggests that the trustworthiness dimensions most affected by compression depend on the specific base model in practice.
\subsection{Alignment-Aware Compression as Existence Proof}
\label{sec:dpo}

The regressions documented above raise a natural question: is it possible to compress reasoning chains without sacrificing core trustworthiness properties?
To show feasibility, we construct alignment-aware DPO~\citep{rafailov2023direct} variants that explicitly encode a preference for concise, high-quality responses across three different base models.

\paragraph{Data and training.}
We curate preference pairs from UltraFeedback~\citep{cui2023ultrafeedback} and Stanford Human Preferences~\citep{ethayarajh2022understanding}, retaining pairs where the chosen response is shorter yet of higher quality.
We fine-tune three base models (DeepSeek-R1-Distill-Qwen-7B, Qwen3-8B, and DeepSeek-R1-0528-Qwen3-8B) with LoRA~\citep{hu2022lora} under the DPO objective~\citep{rafailov2023direct}, implemented in LLaMA-Factory~\citep{zheng2024llamafactory}. The training configuration is summarized in Appendix~\ref{app:dpo}.

\paragraph{Results.}
On Qwen3-8B, the DPO variant achieves 94.8\% on GSM8K and 63.4\% on MATH-500, closely matching the uncompressed baseline at 94.5\% and 63.0\%, while reducing CoT length by 19.3\%.
On the harder AIME24, AIME25, and GPQA-Diamond evaluations in Appendix~\ref{app:hard_reasoning}, DPO changes accuracy by $-3.3$, $-3.3$, and $+6.0$ points, respectively. Its mean CoT token count increases on both AIME sets and decreases slightly on GPQA-Diamond, showing that this variant does not mechanically shorten every reasoning trace.
On trustworthiness~(Table~\ref{tab:main}), HarmBench drops by only 3.8 percentage points compared to $-$10 for L1, AgentHarm remains virtually unchanged at 55.4\%, and MMLU-ProX and FaithEval remain within 1 percentage point of the baseline.
A similar pattern appears in the two additional evaluated base models: on DeepSeek-R1-Distill-Qwen-7B, DPO improves HarmBench while maintaining multilingual and hallucination performance. On DeepSeek-R1-0528-Qwen3-8B, DPO achieves the highest safety score in its group with FaithEval also improving.
These variants are intentionally simple. They demonstrate that preference-guided compression can shorten CoT while limiting trustworthiness loss, but they do not separate the effects of preference-data construction from those of the DPO objective.

\section{Discussion}
\label{sec:discussion}

\subsection{Why does compression affect trustworthiness?}
We conjecture two possible mechanisms.
First, safety alignment from RLHF is encoded in the parameter space that compression modifies, so distillation may partially overwrite these representations.
This echoes findings by \citet{qi2023fine}, who show that even benign fine-tuning can compromise safety alignment by perturbing a small number of critical parameter directions.
\citet{hong2024decoding} similarly observe that compression techniques vary in how well they preserve alignment, with pruning-based methods often causing more damage than quantization-based ones.
Second, the reasoning chain may serve as a deliberation buffer for safety-relevant considerations, and aggressive truncation may remove this buffer.
For instance, a model that generates intermediate tokens reflecting on potential harm would lose this self-revision capacity if those tokens are suppressed.
Validating these hypotheses is a direction for future work, particularly for understanding the observed regressions.

\subsection{Practical and reporting implications}
Our findings have implications for both the deployment and the reporting of compressed reasoning models.
For practitioners, standard reasoning benchmarks alone are insufficient for evaluating compressed models. Safety, hallucination resistance, and multilingual robustness should each be re-evaluated, even when task accuracy appears preserved.
Because different compression methods produce distinct degradation profiles (Section~\ref{sec:patterns}), method selection should be guided by the properties most critical to the target deployment setting.
Our DPO results (Section~\ref{sec:dpo}) further suggest that incorporating objectives that preserve alignment into the compression process can reduce loss on these properties.
For reporting, we recommend that compression papers present both absolute within-family deltas and per-dimension normalized efficiency scores.
Absolute deltas retain the practical meaning of performance changes relative to a matched baseline, while normalized scores support cross-base synthesis without obscuring safety regressions behind gains on other dimensions.
Adopted together, these practices would give a more complete picture of compressed models for both reliable deployment and rigorous scientific comparison.

\subsection{Limitations}
Our study is conducted under a reproducible, baseline-matched protocol. To preserve comparability across model families, we use the same decoding and judging pipeline throughout and include only methods that could be evaluated reliably under this protocol at submission time. This necessarily limits the scope of evaluated techniques. Although human and second-judge audits on the subset show substantial agreement (Appendix~\ref{app:judge_audit}), the full safety evaluation still uses Llama Guard as its primary judge, and item-level validation remains limited to that subset. Our conclusions should therefore be read as recurring empirical patterns observed under a controlled evaluation setup, not as an exhaustive account of the full CoT compression literature. Broader method coverage, larger human audits, and additional independent judges would further strengthen the robustness, generality, and external validity of these patterns.

\section{Conclusion}
\label{sec:conclusion}

Our systematic study reveals that chain-of-thought compression frequently introduces trustworthiness regressions that standard accuracy benchmarks do not reliably detect.
Across evaluations from 1.5B to 32B parameters, with the Qwen3-8B family providing the broadest within-family comparison breadth, three findings emerge:
(1)~compression can degrade safety by up to 10 percentage points while simultaneously improving other trustworthiness dimensions;
(2)~different compression methods produce qualitatively distinct degradation profiles on the same base model;
and (3)~na\"ive scalar efficiency metrics can mask safety regressions that per-dimension analysis clearly reveals.
Consistent trends across multiple base models suggest that these patterns are not isolated to a single model, though establishing such broader claims still requires wider and more balanced method coverage.

CoT compression should therefore optimize not just efficiency but also trustworthiness. Our preference-based variants, implemented with DPO across three base models, provide feasibility evidence without isolating DPO's effect. Future evaluations should report accuracy, token savings, and the preservation of core trustworthiness properties.

\section*{Ethics Statement}
This work evaluates the trustworthiness of LLM compression methods using harmful prompts from established benchmarks, including HarmBench and AgentHarm.
We did not create new harmful benchmark prompts, and all agentic benchmark tasks were executed in a sandboxed environment.
Our findings are intended to inform safer deployment practices for compressed reasoning models.

\bibliography{references}
\bibliographystyle{colm2026_conference}

\clearpage

\appendix
\section{Method Provenance}
\label{app:methods}
Table~\ref{tab:methods} summarizes the compression methods included in our evaluation, together with their paradigms, supported base models, and artifact provenance.

\begin{table}[H]
\centering
\small
\vspace{0.5em}

\begin{tabular}{@{}llll@{}}
\toprule
\textbf{Method} & \textbf{Paradigm} & \textbf{Base Models} & \textbf{Weights} \\
\midrule
L1  & Distillation (RL) & 1.5B, 7B, 8B & Author \\
BeConcise & Prompt eng. & 1.5B, 7B, 8B & N/A \\
Thinkless & RL & 1.5B & Author \\
TokenSkip & Training-free & 8B & Reprod. \\
VeriThinker & Distillation & 7B & Author \\
DPO & Preference-based & 7B, 8B & Ours \\
O1-Pruned & Pruning & 32B  & Author \\
\bottomrule
\end{tabular}
\caption{Compression methods evaluated in this study. ''Author'' indicates official pretrained weights, ''Reprod.'' indicates our reproduction.}
\label{tab:methods}

\end{table}

\section{Method Coverage and Inclusion Criteria}
\label{app:coverage}

We did not attempt to enumerate the literature solely by citation count.
Instead, we scoped our evaluation to the subset of methods that could be assessed credibly under a common trustworthiness protocol at submission time.
Our inclusion policy prioritized artifact availability and reproducibility: a method was included if it had usable public checkpoints on a supported base model, or if public code or configuration enabled a stable reproduction, or if it was a training-free intervention directly applicable to a supported base model.
This policy is intended to reduce hidden degrees of freedom from ad hoc re-implementations while making the evaluation reproducible for future work.

\begin{table}[H]
\centering
\small

\vspace{0.5em}
\begin{tabular}{@{}>{\raggedright\arraybackslash}p{0.24\linewidth}>{\raggedright\arraybackslash}p{0.14\linewidth}>{\raggedright\arraybackslash}p{0.5\linewidth}@{}}
\toprule
\textbf{Category} & \textbf{Status} & \textbf{Criterion / reason} \\
\midrule
Official checkpoints on supported base models & Included & Evaluated directly with author-released weights (e.g., L1, Thinkless, VeriThinker, O1-Pruned). \\
Methods reproducible from public code but without released weights & Included & Reproduced when the public implementation was sufficiently specified for a stable run in our evaluation stack (e.g., TokenSkip, model merging). \\
Training-free interventions applicable to supported bases & Included & No retraining required, directly evaluated as inference-time baselines (e.g., BeConcise). \\
Our preference-based compression variant & Included & Trained and evaluated by us on DeepSeek-R1-Distill-Qwen-7B, Qwen3-8B, and DeepSeek-R1-0528-Qwen3-8B to provide feasibility evidence for preference-based compression. \\
Methods without usable public artifacts & Excluded & Omitted when neither runnable checkpoints nor sufficiently detailed public artifacts were available at submission time. \\
Methods with insufficient reproduction detail or incompatible evaluation path & Excluded & Omitted when available materials did not support a stable and auditable reproduction under our common protocol. \\
\bottomrule
\end{tabular}
\caption{Coverage policy for methods considered in this study. Our evaluation targets the currently available and reproducible subset of CoT compression methods rather than the full set of papers in the literature.}
\label{tab:coverage}
\end{table}

\vspace{10pt}

\noindent
Under this policy, our results should be interpreted as a systematic evaluation of the currently available and reproducible CoT compression methods.
We do not claim exhaustive coverage of every published compression proposal.

\section{DPO Training Configuration}
\label{app:dpo}

The DPO variants are trained using LLaMA-Factory with the following configuration:
LoRA rank 16, scaling factor 32, dropout 0.05, DPO loss coefficient 0.1 with sigmoid loss, learning rate $1 \times 10^{-5}$, cosine schedule with 3\% warmup, batch size 2 $\times$ 8 gradient accumulation steps, FP16 precision, 1 epoch, cutoff length 2048 tokens.
We apply this identical configuration to three base models: DeepSeek-R1-Distill-Qwen-7B, Qwen3-8B, and DeepSeek-R1-0528-Qwen3-8B.
Training data consists of filtered preference pairs from UltraFeedback and SHP where chosen responses are shorter yet higher quality than rejected responses.

\section{Safety Judge Audits}
\label{app:judge_audit}

We check the reliability of the Llama-Guard automatic safety judge on a sampled subset of 513 judgments. The same subset is used for two comparisons: one against human annotations and one against \texttt{deepseek-v4-flash}, which applies the same safety rubric. Table~\ref{tab:judge_audits} reports the complete aggregate confusion matrices supplied by these audits.

\begin{table}[H]
\centering
\small
\begin{tabular}{@{}lcc@{}}
\toprule
\multicolumn{3}{c}{\textbf{Llama Guard vs. human annotations}} \\
 & \textbf{Human unsafe} & \textbf{Human safe} \\
\midrule
Llama Guard unsafe & 175 & 30 \\
Llama Guard safe & 20 & 288 \\
\midrule
\multicolumn{3}{c}{\textbf{Llama Guard vs. \texttt{deepseek-v4-flash}}} \\
 & \textbf{DeepSeek unsafe} & \textbf{DeepSeek safe} \\
\midrule
Llama Guard unsafe & 178 & 27 \\
Llama Guard safe & 18 & 290 \\
\bottomrule
\end{tabular}
\caption{Confusion matrices on the same 513-case safety-judgment subset. Rows are Llama-Guard labels; columns are the comparison labels.}
\label{tab:judge_audits}
\end{table}

Llama Guard agrees with the human annotations on 463/513 cases (90.25\%), with Cohen's $\kappa=0.795$. It agrees with \texttt{deepseek-v4-flash} on 468/513 cases (91.23\%), with $\kappa=0.816$. These results support the reliability and judge-choice stability of the reported safety labels on this sampled subset; they do not establish agreement for every item in the full evaluation.

\section{Hard Reasoning Evaluation}
\label{app:hard_reasoning}

Table~\ref{tab:hard_reasoning} extends the Qwen3-8B DPO evaluation beyond GSM8K and MATH-500. The mixed CoT-length changes show that the variant does not act as a fixed length cap: it uses longer traces on the two AIME sets in this evaluation and a slightly shorter trace on GPQA-Diamond.

\begin{table}[H]
\centering
\footnotesize
\setlength{\tabcolsep}{2.5pt}
\begin{tabular}{@{}l cc cc cc@{}}
\toprule
& \multicolumn{2}{c}{\textbf{AIME24}} & \multicolumn{2}{c}{\textbf{AIME25}} & \multicolumn{2}{c}{\textbf{GPQA-Diamond}} \\
\cmidrule(lr){2-3} \cmidrule(lr){4-5} \cmidrule(lr){6-7}
\textbf{Method} & \textbf{Acc.} & \textbf{CoT tokens} & \textbf{Acc.} & \textbf{CoT tokens} & \textbf{Acc.} & \textbf{CoT tokens} \\
\midrule
Baseline & 73.3 & 11236.37 & 60.0 & 13822.08 & 57.1 & 7478.44 \\
DPO & 70.0 ($-3.3$) & 11738.69 ($+4.5\%$) & 56.7 ($-3.3$) & 14745.36 ($+6.7\%$) & 63.1 ($+6.0$) & 7290.50 ($-2.5\%$) \\
\bottomrule
\end{tabular}
\caption{Hard reasoning accuracy (\%) and mean CoT token count for Qwen3-8B. Parentheses in the DPO row give the accuracy-point change and relative CoT-token change from the baseline.}
\label{tab:hard_reasoning}
\end{table}

\section{Sensitivity Analysis of \texorpdfstring{$E$}{E}}
\label{app:sensitivity}

To verify that our conclusions are not artifacts of the specific exponent choices $\alpha = 2$, $\beta = 0.5$, we recompute $E$ for all Qwen3-8B methods under a grid of $\alpha \in \{1, 2, 3\}$ and $\beta \in \{0.3, 0.5, 0.7\}$ (9 configurations total).

\begin{table}[H]
\centering
\scriptsize
\setlength{\tabcolsep}{2.5pt}
\begin{tabular}{@{}l ccc ccc ccc@{}}
\toprule
 & \multicolumn{3}{c}{$\alpha=1$} & \multicolumn{3}{c}{$\alpha=2$} & \multicolumn{3}{c}{$\alpha=3$} \\
\cmidrule(lr){2-4} \cmidrule(lr){5-7} \cmidrule(lr){8-10}
\textbf{Method} & $\beta\!=\!.3$ & $\beta\!=\!.5$ & $\beta\!=\!.7$ & $\beta\!=\!.3$ & $\beta\!=\!.5$ & $\beta\!=\!.7$ & $\beta\!=\!.3$ & $\beta\!=\!.5$ & $\beta\!=\!.7$ \\
\midrule
L1-Exact & 0.910 & 0.949 & 0.990 & 0.778 & 0.811 & 0.846 & 0.665 & 0.693 & 0.723 \\
L1-Max & 0.923 & 0.980 & 1.040 & 0.778 & 0.826 & 0.877 & 0.656 & 0.696 & 0.739 \\
BeConcise & 0.848 & 0.806 & 0.766 & 0.777 & 0.738 & 0.701 & 0.711 & 0.676 & 0.642 \\
TokenSkip & 1.042 & 1.108 & 1.179 & 0.989 & 1.052 & 1.119 & 0.938 & 0.999 & 1.063 \\
DPO & 0.959 & 0.969 & 0.979 & 0.906 & 0.916 & 0.925 & 0.856 & 0.865 & 0.874 \\
\bottomrule
\end{tabular}
\caption{Sensitivity of $E_{\text{HB}}$ (HarmBench) on Qwen3-8B to $(\alpha, \beta)$ variation.}
\label{tab:sensitivity_hb}
\end{table}

\begin{table}[H]
\centering
\scriptsize
\setlength{\tabcolsep}{2.5pt}
\begin{tabular}{@{}l ccc ccc ccc@{}}
\toprule
 & \multicolumn{3}{c}{$\alpha=1$} & \multicolumn{3}{c}{$\alpha=2$} & \multicolumn{3}{c}{$\alpha=3$} \\
\cmidrule(lr){2-4} \cmidrule(lr){5-7} \cmidrule(lr){8-10}
\textbf{Method} & $\beta\!=\!.3$ & $\beta\!=\!.5$ & $\beta\!=\!.7$ & $\beta\!=\!.3$ & $\beta\!=\!.5$ & $\beta\!=\!.7$ & $\beta\!=\!.3$ & $\beta\!=\!.5$ & $\beta\!=\!.7$ \\
\midrule
L1-Exact & 1.210 & 1.343 & 1.490 & 1.252 & 1.390 & 1.542 & 1.296 & 1.438 & 1.596 \\
L1-Max & 1.220 & 1.375 & 1.549 & 1.244 & 1.402 & 1.580 & 1.269 & 1.430 & 1.611 \\
BeConcise & 1.080 & 1.102 & 1.124 & 1.130 & 1.154 & 1.177 & 1.183 & 1.208 & 1.233 \\
TokenSkip & 0.972 & 1.012 & 1.054 & 0.890 & 0.926 & 0.964 & 0.814 & 0.848 & 0.882 \\
DPO & 1.088 & 1.154 & 1.224 & 1.083 & 1.148 & 1.218 & 1.078 & 1.143 & 1.213 \\
\bottomrule
\end{tabular}
\caption{Sensitivity of $E_{\text{MX}}$ (MMLU-ProX) on Qwen3-8B to $(\alpha, \beta)$ variation.}
\label{tab:sensitivity_mx}
\end{table}

\begin{table}[H]
\centering
\scriptsize
\setlength{\tabcolsep}{2.5pt}
\begin{tabular}{@{}l ccc ccc ccc@{}}
\toprule
 & \multicolumn{3}{c}{$\alpha=1$} & \multicolumn{3}{c}{$\alpha=2$} & \multicolumn{3}{c}{$\alpha=3$} \\
\cmidrule(lr){2-4} \cmidrule(lr){5-7} \cmidrule(lr){8-10}
\textbf{Method} & $\beta\!=\!.3$ & $\beta\!=\!.5$ & $\beta\!=\!.7$ & $\beta\!=\!.3$ & $\beta\!=\!.5$ & $\beta\!=\!.7$ & $\beta\!=\!.3$ & $\beta\!=\!.5$ & $\beta\!=\!.7$ \\
\midrule
L1-Exact & 1.153 & 1.234 & 1.320 & 1.202 & 1.286 & 1.376 & 1.253 & 1.341 & 1.434 \\
L1-Max & 1.158 & 1.273 & 1.400 & 1.165 & 1.280 & 1.407 & 1.171 & 1.287 & 1.414 \\
BeConcise & 0.971 & 0.973 & 0.975 & 0.940 & 0.942 & 0.944 & 0.911 & 0.912 & 0.914 \\
TokenSkip & 1.051 & 1.094 & 1.138 & 1.040 & 1.082 & 1.126 & 1.029 & 1.071 & 1.115 \\
DPO & 1.034 & 1.064 & 1.096 & 1.023 & 1.053 & 1.084 & 1.012 & 1.042 & 1.073 \\
\bottomrule
\end{tabular}
\caption{Sensitivity of $E_{\text{FE}}$ (FaithEval) on Qwen3-8B to $(\alpha, \beta)$ variation.}
\label{tab:sensitivity_fe}
\end{table}

\begin{table}[H]
\centering
\small
\begin{tabular}{@{}l ccc@{}}
\toprule
\textbf{Benchmark} & \textbf{Direction stability} & \textbf{Pairwise stability} & \textbf{Mean Kendall $\tau$} \\
\midrule
HarmBench & 41/45 (91.1\%) & 83/90 (92.2\%) & 0.844 \\
MMLU-ProX & 43/45 (95.6\%) & 83/90 (92.2\%) & 0.844 \\
FaithEval & 45/45 (100.0\%) & 86/90 (95.6\%) & 0.911 \\
\midrule
Overall & 129/135 (95.6\%) & 252/270 (93.3\%) & 0.867 \\
\bottomrule
\end{tabular}
\caption{Stability relative to the default $(\alpha,\beta)=(2,0.5)$. Direction denotes whether $E$ is above or below 1; pairwise stability counts the 10 method pairs per benchmark and configuration.}
\label{tab:sensitivity_summary}
\end{table}

Across the 27 benchmark--configuration combinations, the sign of $E-1$ is unchanged in 129/135 method cases (95.6\%), and pairwise method order is unchanged in 252/270 comparisons (93.3\%); the mean Kendall $\tau$ is 0.867 (Table~\ref{tab:sensitivity_summary}). The central L1 pattern is even more stable: the two L1 variants yield $E_{\text{HB}}<1$ and $E_{\text{MX}},E_{\text{FE}}>1$ in 53/54 cases, with the sole exception being L1-Max at $(\alpha,\beta)=(1,0.7)$ on HarmBench. Thus, exponent choice can affect borderline method orderings, but it does not drive the main dimension-specific L1 conclusion.

\section{Bootstrap Stability of Pareto Frontiers}
\label{app:pareto_bootstrap}

We assess sample sensitivity using 10,000 paired bootstrap replicates per benchmark. In each replicate, test examples are sampled with replacement; every method is evaluated on the same resampled examples, after which its benchmark metric and mean CoT length are recomputed and the original Pareto rule is reapplied. Table~\ref{tab:pareto_bootstrap} reports both exact-set recovery and the inclusion frequency of original high-frequency frontier members.

\begin{table}[H]
\centering
\small
\begin{tabular}{@{}l c p{0.52\textwidth}@{}}
\toprule
\textbf{Benchmark} & \textbf{Exact set} & \textbf{Stable original members (inclusion frequency)} \\
\midrule
HarmBench & 30.8\% & Baseline 91.2\%; TokenSkip 99.8\% \\
AgentHarm & 67.8\% & BeConcise 77.7\%; TokenSkip 100.0\% \\
MMLU-ProX & 76.6\% & L1-Exact 91.7\%; L1-Max 100.0\%; BeConcise 84.7\% \\
FaithEval & 95.3\% & L1-Exact 97.5\%; L1-Max 100.0\% \\
\bottomrule
\end{tabular}
\caption{Pareto-frontier stability over 10,000 paired bootstrap replicates. Exact set is the frequency with which the complete original frontier-member set is recovered.}
\label{tab:pareto_bootstrap}
\end{table}

The lower exact-set recovery on HarmBench reflects boundary uncertainty from nearby trade-off points occasionally entering the frontier, rather than displacement of its original Baseline and TokenSkip anchors, which remain members in 91.2\% and 99.8\% of replicates. The other benchmarks show higher exact-set recovery, and their listed original members also recur frequently. These results support the qualitative cross-dimension comparison while qualifying claims about the exact membership of any one empirical frontier.

\section{Full Pareto Analysis}
\label{app:pareto}

\begin{figure}[H]
\centering
\includegraphics[width=\textwidth]{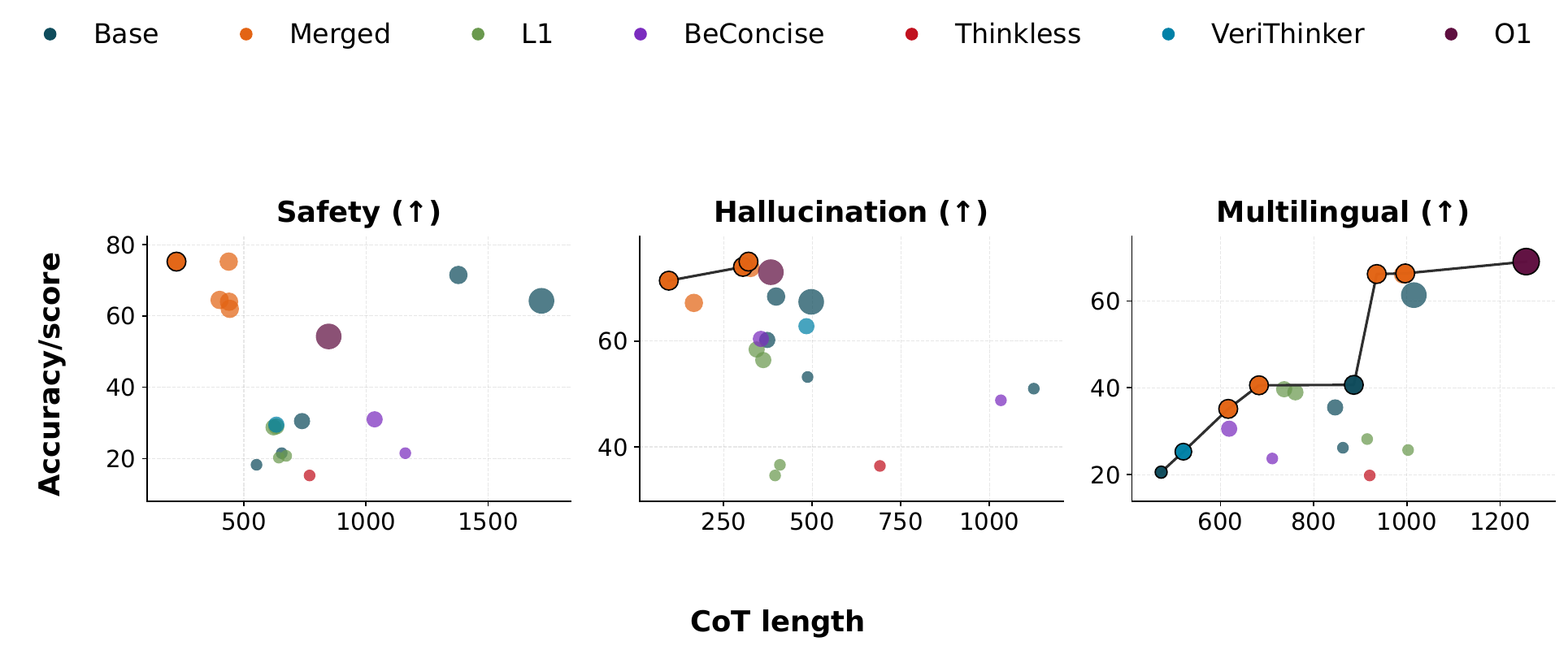}
\caption{Trustworthiness--length Pareto frontiers across all model groups. Colors denote compression methods, marker size denotes model scale. The black polyline marks the Pareto frontier in each panel.}
\label{fig:pareto_all}
\end{figure}

The main text (Section~\ref{sec:pareto}) presents Pareto frontiers for the Qwen3-8B family, which provides the most complete method coverage.
Figure~\ref{fig:pareto_all} extends the same analysis to all model families.
The cross-family plot shows that frontier membership depends heavily on the base model: merged variants dominate on HarmBench, L1 variants lead on MMLU-ProX and FaithEval, and no single method occupies the frontier across all dimensions.

\end{document}